\documentclass[letterpaper, 10 pt, conference]{ieeeconf}

\IEEEoverridecommandlockouts
\usepackage{etoolbox} 
\usepackage[utf8]{inputenc} 
\usepackage[T1]{fontenc}    
\usepackage{fonttable}
\usepackage[english]{babel} 
\usepackage[protrusion=true,expansion=true]{microtype}
\usepackage{comment}
\usepackage{cancel}
\usepackage{csquotes}
\usepackage{listings}
\usepackage{nameref}
\usepackage{paralist}
\usepackage{xspace}
\usepackage{setspace}
\usepackage{makeidx}
\usepackage{pdfpages}

\usepackage{amsmath,amssymb} 
\usepackage{amsfonts}       
\usepackage{mathtools}
\usepackage{array} 
\usepackage{nicefrac}       
\usepackage{breqn}          
\usepackage{textcomp}
\usepackage{bm} 
\usepackage{pifont}
\usepackage[
  separate-uncertainty = true,
  multi-part-units = repeat
]{siunitx}
\usepackage{mathrsfs}

\usepackage{graphicx}
\usepackage{wrapfig}
\usepackage{adjustbox} 
\usepackage{epsfig}

\usepackage{float}
\usepackage{capt-of}
\usepackage{lscape}      

\usepackage{tikz}  
\usepackage{makecell}
\usepackage{overpic}

\newsavebox{\subfigbox}

\usepackage{algorithm}
\usepackage[noend]{algpseudocode}

\usepackage{array} 
\usepackage{tabularx}
\usepackage{multirow}
\usepackage{multicol}
\usepackage{booktabs}   
\usepackage{tablefootnote}

\usepackage{paralist}
\let\labelindent\relax 
\usepackage{enumitem}

\usepackage{url}            
\usepackage[table]{xcolor}
\usepackage{colortbl}
\usepackage{soul}

\PassOptionsToPackage{capitalize, noabbrev}{cleveref}

\makeatletter
\let\NAT@parse\undefined
\makeatother
\usepackage{cite}
\usepackage{hyperref}
\hypersetup{breaklinks=true,colorlinks,urlcolor=blue,citecolor=blue,linkcolor=blue}

\usepackage{blindtext}
\usepackage{bbm}

\usepackage{lipsum}

\makeatletter
\def\fnum@table{Table~\thetable}
\long\def\@IEEEcaptionbody#1#2{%
  \setbox\@tempboxa\hbox{\footnotesize\textbf{#1:}~~#2}%
  \ifdim\wd\@tempboxa>\hsize
    \setbox\@tempboxa\hbox{\footnotesize\textbf{#1:}~~}%
    \parbox[t]{\hsize}{\footnotesize\noindent\unhbox\@tempboxa#2}%
  \else
    \hbox to\hsize{\footnotesize\box\@tempboxa\hfil}%
  \fi}
\long\def\@makecaption#1#2{%
  \ifx\@captype\@IEEEtablestring
    \@IEEEcaptionbody{#1}{#2}%
    \@IEEEtablecaptionsepspace
  \else
    \@IEEEfigurecaptionsepspace
    \@IEEEcaptionbody{#1}{#2}%
  \fi}
\makeatother

\definecolor{color1}{rgb}{.6,.4,.05}
\definecolor{color2}{rgb}{0,.7,.7}
\definecolor{color3}{rgb}{0.35,0.75,0.0}
\definecolor{color4}{rgb}{0.4,0.8,0}
\definecolor{color5}{rgb}{0.5,0.0,0.5}
\definecolor{revision_color}{rgb}{1,0,0}

\newcommand{\modelNameHeight}{1.6ex}
\newcommand{\modelLogo}{\protect\raisebox{-0.18ex}{\includegraphics[height=\modelNameHeight]{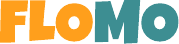}}\xspace}
\newcommand{\modelName}{FloMo\xspace}

\title{\modelLogo{}: 3D Scene Flow as a World Action Model  \\ Intermediate for Learning from Human Video}

\author{%
  Jeremy A.~Collins$^{*}$, Namra Patel$^{*}$, Ayush Agarwal$^{*}$, Shitij
Govil$^{*}$, and Animesh Garg \\ 
  Georgia Institute of Technology%
  \thanks{$^{*}$Equal contribution.}%
  \thanks{Correspondence: \texttt{sgovil9@gatech.edu}}%
}

\begin{document}
\maketitle
\bstctlcite{IEEEexample:BSTcontrol} 


\begin{abstract}
Pretrained video generation models distill internet-scale motion priors into robot policies, but full video is a redundant signal for action prediction. We argue that dense 3D motion is the medium through which video pretraining best transfers to robot actions, and that predicting motion rather than video affords better sample-efficiency for action prediction. \modelName{} is a 5B video-generation model fine-tuned to predict 3D scene-flow and action tokens. Rendering scene flow as RGB and encoding it into the same latent space as video enables a pretrained video generation backbone to efficiently adapt to predict 3D scene flow. Co-trained on action-free human video and action-labeled robot teleoperation data, our model benefits from human-video pretraining and generalizes to unseen objects, motion primitives, and tasks. Replacing motion prediction with a video prediction objective degrades out-of-distribution generalization, while adding a video prediction target alongside motion lowers policy success rather than improving it. When motion, video, and action are predicted jointly, action tokens attend far more to motion than to video, suggesting that predicting video wastes model capacity. Together, these results indicate that motion effectively mediates the transfer from video pretraining to action. More information can be found at \href{https://flomo-wam.github.io/}{flomo-wam.github.io}.
\end{abstract}


\section{Introduction}
\label{sec:intro}

Pretrained video generation models can be used to distill internet-scale motion priors into robot policies. Transfer from video to action is typically attributed to information overlap between RGB videos and the actions that produce them: a model that predicts the world's next pixels has implicitly learned the dynamics that link action to outcome~\cite{Jang2025DreamGen, Liang2025VideoPolicy, Ye2026DreamZero}. Video, however, is a far higher-entropy signal than the action it implies. Videos that differ in texture, lighting, material, and background depict the same underlying motion and imply the same action, so a world action model (WAM) trained to predict pixels spends capacity on appearance that action prediction never uses~\cite{Wen2024ATM, Collins2025AMPLIFY, Xu2024Im2Flow2Act}.

WAMs vary on how tightly they couple the video backbone to the action head, but they share a prediction target: future pixels. We question this assumption and ask: \textit{which prediction target lets video pretraining transfer to action most directly?} We argue that the answer is 3D motion. Given the current observation, the action that produces a future state depends on how the scene transformed rather than on appearance and texture. Predicted frames that differ only in lighting or background share the same motion and imply the same action. Predicting motion rather than pixels hence targets a lower-entropy signal and incorporates this invariance instead of forcing the model to learn it, leaving the action head a tighter information bottleneck it can decode directly.

Our insight is that a pretrained video VAE already provides a latent space for motion. We represent motion as the 3D displacement of each tracked scene point over time in a canonical frame independent of camera motion, otherwise known as \emph{3D scene flow}, and render this vector field as an RGB image. Rendering scene flow as RGB lets us encode it with a pretrained video VAE, giving scene-flow tokens the same latent representation format as video without needing to train a new tokenizer. This lets us finetune only the backbone to predict motion.

We instantiate this in \modelLogo{}, a Wan~2.2 5B DiT that, given an image and language instruction, flow-matches 3D scene-flow and action tokens, with scene flow encoded into the same video latent space the model was pretrained on. We co-train on action-free human video and action-labeled robot teleoperation data, and find that predicting motion alone improves both in-distribution policy success and out-of-distribution generalization, relative to predicting video or predicting video and motion jointly.

Our contributions are:
\begin{enumerate}
    \item \textbf{Motion as a WAM prediction target (\S\ref{sec:method}).} We argue that WAMs should predict motion rather than video: motion is a lower-complexity, texture-invariant target an action head can decode directly, and the priors of video pretraining are carried by the pretrained backbone, not by the prediction target. We realize this by rendering 3D scene flow as an RGB image and encoding it with the pretrained video VAE, so scene-flow tokens share the video latent space and a single backbone flow-matches motion and action without a new tokenizer.
    \item \textbf{Motion is the most effective prediction target for policy learning (\S\ref{sec:in-dist}, \S\ref{sec:ood}).} Across real-robot manipulation, predicting motion yields higher in-distribution success and out-of-distribution generalization than predicting video or jointly predicting video and motion; adding a video target interferes with action learning rather than complementing it.
    \item \textbf{Generalization from human video (\S\ref{sec:human-video}).} Our model generalizes to objects, motion primitives, and tasks seen only in human data.
\end{enumerate}


\section{Related Work}
\label{sec:related}
\begin{figure*}[!t]
    \centering
    \includegraphics[width=\textwidth]{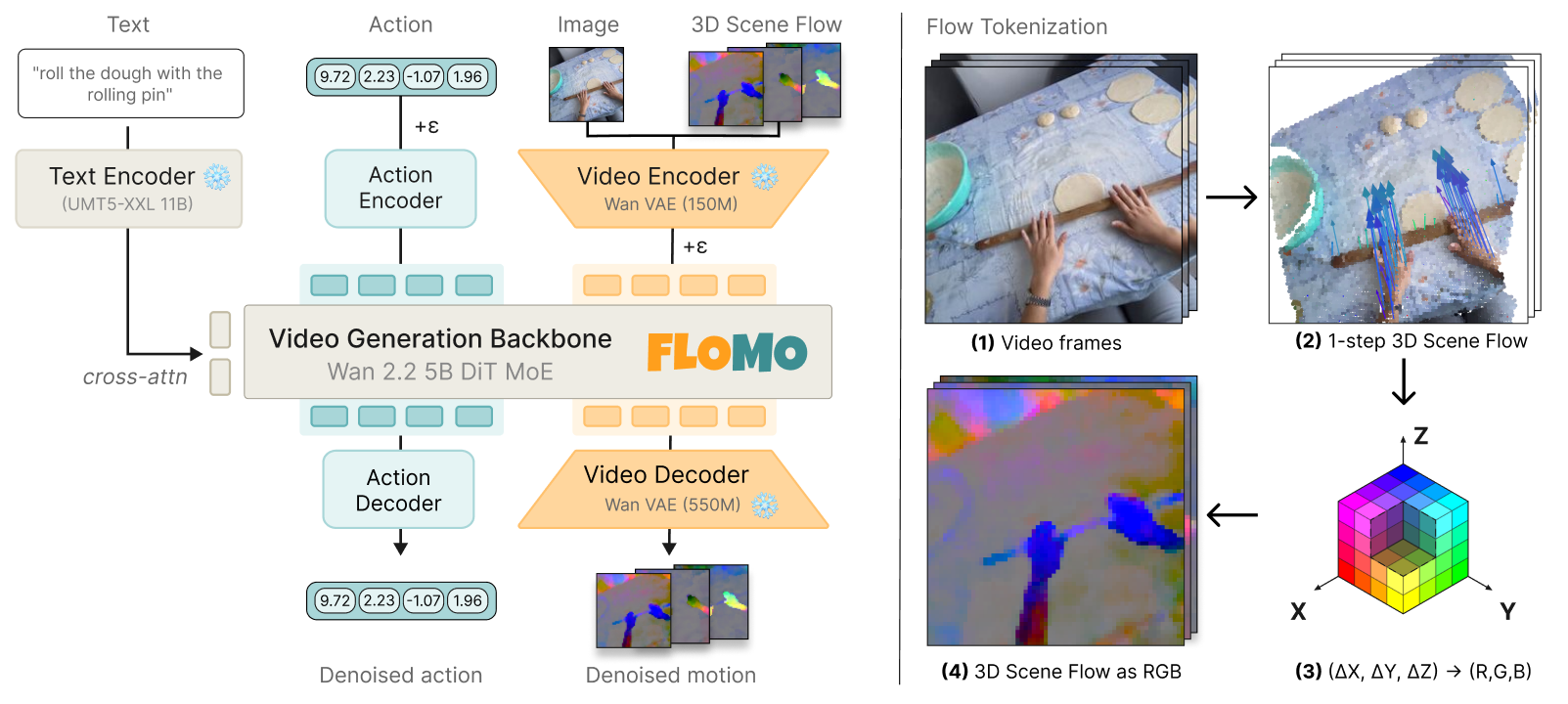}
    \caption{Given an image and text prompt, \modelName{} jointly predicts dense 3D motion (scene flow) and robot actions using a pretrained video generation model using flow-matching. To tokenize scene flow, we transform cumulative 3D velocities into RGB videos that are encoded by the pretrained video backbone's video encoder.}
    \label{fig:figure_arch}
    \vspace{-2mm}
\end{figure*}

\subsection{Learning Robot Policies from Action-Free Videos}
Action-labeled robot demonstrations are expensive, while large-scale action-free videos of humans, robots, and objects are abundant. One line of work uses passive video to pretrain visual representations, rewards, or value functions (\cite{Nair2022R3M}, \cite{Ma2022VIP}, \cite{Bhateja2023VPTR}, \cite{Dashora2025ViVa}). These methods improve data efficiency but do not learn explicit predictive models of future dynamics.

Another line of work learns latent actions or transition variables from unlabeled videos through self-supervised objectives such as future prediction, frame consistency, or reconstruction (\cite{Bruce2024Genie}, \cite{Ye2024LAPA}, \cite{Chen2024Moto}). While these methods reduce reliance on action labels, their latents may encode visual information useful for reconstruction but unnecessary for control.

Finally, work on learning from human videos studies cross-embodiment alignment or explicit retargeting to transfer human demonstrations to robots (\cite{Punamiya2025EgoBridge}, \cite{Kareer2024EgoMimic}, \cite{Kim2025UniSkill}, \cite{Li2024OKAMI}, \cite{Dan2025XSim}). These works focus on bridging the human-robot embodiment gap; our work instead focuses on the prediction target inside of a WAM.

\subsection{Motion and Geometry-Aware Video Prediction}
Raw pixel prediction is an expensive and indirect objective for learning physical dynamics, motivating methods that introduce motion or geometric intermediates. Some use 2D motion as an interface between action-free video and robot actions, predicting future image-point trajectories and conditioning policies on them (\cite{Wen2024ATM}, \cite{Collins2025AMPLIFY}). Others use flow or motion fields as an interface linking video prediction and manipulation (\cite{Yuan2024GeneralFlow}, \cite{Xu2024Im2Flow2Act}, \cite{Lin2024FlowRetrieval}, \cite{Gao2024FLIP}), or separate motion and temporal reasoning from appearance and camera movement in video generation (\cite{Liu2026MoRight}).

Closest to our representation choice are methods that model dynamic scenes with compact trace spaces, 3D scene flow, or point-track representations (\cite{Lee2025TraceGen}, \cite{Guo2025FlowDreamer}, \cite{Hung2026ThreePoinTr}). These works show that 3D motion fields are a natural substrate for future dynamics prediction. Rather than building a bespoke latent space for motion, we render scene flow into a pretrained video-generation latent space.

\subsection{World Action Models}
Pretrained video-generation models provide motion priors for robot learning because they are trained on large-scale videos of physical interactions and object motion. Recent work uses these models for control by coupling future video prediction with action generation. One family generates task-conditioned robot videos and then recovers or co-generates actions through inverse dynamics models or latent actions (\cite{Du2023UniPi}, \cite{Ko2024AVDC}, \cite{Jang2025DreamGen}, \cite{Liang2025VideoPolicy}). These approaches preserve the pretrained video model, but deployment is expensive and performance depends heavily on video fidelity. A second family avoids explicit video rollout at test time by extracting intermediate latents from a pretrained video backbone and training an action head on top (\cite{Wen2024VidMan}, \cite{Hu2024VPP}, \cite{Pai2025MimicVideo}). This reduces inference cost, but action prediction remains limited by the quality of the extracted video features.

More tightly coupled world action models jointly model visual features and actions with shared or entangled architectures (\cite{Zhu2025UWM}, \cite{Li2025UVA}, \cite{Kim2026CosmosPolicy}, \cite{Ye2026DreamZero}, \cite{Yuan2026FastWAM}). Our work studies a complementary axis: what future state should the world model predict? Instead of pixels, video latents, or action tokens, we predict 3D scene flow, discarding texture, lighting, and material details that are irrelevant for inverse dynamics.


\section{Method}
\label{sec:method}

Existing WAMs predict future video jointly with actions, forcing the backbone to model details that low-level control doesn't require. \modelLogo{} predicts 3D scene flow in place of video. We predict within the latent space of a pretrained video generator (Wan~2.2), so the model inherits large-scale video priors through its backbone. Our model is co-trained on egocentric human video and robot teleoperation, predicting both scene flow and actions from an initial observation and a language instruction. Figure~\ref{fig:figure_arch} gives an overview of the architecture.

\subsection{Problem Formulation}
\label{sec:problem}
We assume access to two types of datasets: an \emph{egocentric video dataset} $\mathcal{V} = \{(\mathbf{o}_{0:T-1},\, \ell)\}$ of RGB observation sequences $\mathbf{o}_{0:T-1} \in \mathbb{R}^{T \times H \times W \times 3}$ ($T = 16$) paired with language instructions $\ell$, and a \emph{robot interaction dataset} $\mathcal{R} = \{(\mathbf{o}_{0:T-1},\, \ell,\, \mathbf{a})\}$ that additionally provides an action chunk $\mathbf{a} = (\mathbf{a}_0, \dots, \mathbf{a}_{T-1}) \in \mathbb{R}^{T \times d_a}$ over a $d_a$-dimensional action space. For all clips in both datasets, we extract a 3D scene-flow sequence $\mathbf{F}_{0:T-1}$ in an offline preprocessing step (Sec.~\ref{sec:flow}). Given an initial observation $\mathbf{o}_0$ and instruction $\ell$, our model predicts future scene flow and actions; the full joint distribution our architecture represents is:
\begin{equation}
p\!\left(\mathbf{F}_{0:T-1},\; \mathbf{a} \;\middle|\; \mathbf{o}_0,\, \ell\right),
\label{eq:joint}
\end{equation}

\subsection{3D Scene Flow as the Video-Action Intermediate}
\label{sec:flow}

The action producing a future state depends on how the scene moves, not on how it looks, motivating motion rather than pixels as the prediction target. 2D optical flow captures motion but conflates camera ego-motion with task-relevant motion. We instead predict 3D scene flow in a canonical frame, which isolates the geometric transition between states while discarding appearance and camera motion.

\textbf{Scene-flow extraction and canonical frame.}
We extract 3D point trajectories in an offline preprocessing step using SpatialTrackerV2~\cite{Xiao2025SpatialTrackerV2}, an off-the-shelf 3D point tracker. For each clip in $\mathcal{V}$ and $\mathcal{R}$, we slide a window of $T = 16$ frames and place $Q = 4096$ query points on a $64 \times 64$ grid covering the first frame. The tracker back-projects these queries to 3D and propagates them across the window, yielding trajectories $\{\mathbf{p}_i(u,v)\}_{i=0}^{T-1}$. We express every trajectory in the camera frame of the window's first (conditioning) timestep, so the coordinate origin is the current camera's focal point. This makes the representation invariant to camera ego-motion and yields a motion field that primarily reflects object and hand movement. Extraction runs at ${\approx}2.4$\,s per $16$-frame window on a single NVIDIA L40S GPU.

\textbf{Cumulative scene flow.}
Rather than supervising on instantaneous per-frame velocities, we predict \emph{cumulative} scene flow, defined as the total 3D displacement of each point from the first frame to frame $i$:
\vspace{-2mm}
\begin{equation}
    \mathbf{F}_i(u,v) \;=\; \mathbf{p}_i(u,v) \;-\; \mathbf{p}_0(u,v),
    \label{eq:cumflow}
\end{equation}
so $\mathbf{F}_0 = \mathbf{0}$ by construction. We find cumulative scene flow to be a smoother target than instantaneous scene flow: it reduces the per-frame tracking noise that dominates the supervision signal, letting fine-grained motion like fingertip and object movement be predicted accurately.

\subsection{Tokenized Scene Flow}
\label{sec:wan_vae}

We render the cumulative scene flow $\mathbf{F}_{0:T-1}$ as an RGB video $\mathbf{F}^{\mathrm{rgb}}$ by clipping each displacement channel to its global per-dataset 1st/99th percentile range and rescaling to $[0,1]$, then treat it exactly like an RGB video. We encode it with the frozen Wan~2.2-TI2V VAE encoder $\mathcal{E}$:
\begin{equation}
    \mathbf{z}^f \;=\; \mathcal{E}\!\left(\mathbf{F}^{\mathrm{rgb}}\right) \;\in\; \mathbb{R}^{T_z \times c \times H_z \times W_z},
\end{equation}
where $T_z$ is the number of latent frames (temporal stride $s_T = 4$, with the first frame encoded on its own to match the TI2V backbone), $c = 48$ is the VAE latent channel count, and $H_z \times W_z$ are the latent spatial dimensions (spatial stride $16$). Representing scene flow as an RGB video lets us reuse the pretrained VAE without modification, so scene flow and video latents share similar distributions and the backbone does not significantly diverge from its video pretraining.


\subsection{Architecture}
\label{sec:architecture}

\modelLogo{} is instantiated from the pretrained Wan~2.2 5B Text-Image-to-Video DiT~\cite{Wan2025}, extended with lightweight modality-specific encoder and decoder heads. The current observation is encoded by the frozen VAE, $\mathbf{z}^o = \mathcal{E}(\mathbf{o}_0)$, forming a clean conditioning prefix, while the frozen umT5-XXL text encoding
  $\boldsymbol{\tau} = \mathrm{umT5}(\ell)$~\cite{Chung2023UniMax} conditions the backbone through cross-attention. The cumulative scene-flow latents $\mathbf{z}^f$ and action tokens are the prediction targets; future RGB latents $\mathbf{z}^v = \mathcal{E}(\mathbf{o}_{1:T-1})$ are an additional target used only in video-prediction ablations. For the actions, flow matching is performed on the normalized $d_a$-dimensional action space. At noise level $t$, each noised action $a_{t,i} \in \mathbb{R}^{d_a}$ is mapped to a DiT token by a learned linear projection $\mathbb{R}^{d_a} \rightarrow \mathbb{R}^d$. 
  
  All target tokens are concatenated along the sequence dimension and jointly denoised by the shared DiT backbone. We fine-tune the backbone using LoRA~\cite{Hu2022LoRA} ($r=64$, $\alpha=128$, dropout $0$), injecting adapters into the query/key/value/output projections of both self- and cross-attention and into both feed-forward layers of all $30$ blocks; the Wan VAE and text encoder stay frozen. This leaves ${\approx}161$M trainable parameters on top of the frozen 5B backbone. We use bidirectional (full) attention, as in Wan, so every token attends to every other. Each prediction head can be independently activated or dropped per sample via loss masking, so the same backbone supports any subset of output modalities. By default, our model predicts scene flow and actions; our ablations reuse the same flexible architecture to additionally predict future video, or to predict video and actions without scene flow.

\subsection{Training Objective}
\label{sec:objective}

We adopt a flow-matching objective~\cite{Lipman2022FlowMatching} over all prediction targets. For any clean target $\mathbf{x}_0 \in \{\mathbf{z}^f, \mathbf{a}, \mathbf{z}^v\}$, we form the noisy interpolant at a single noise level $t \sim \mathcal{U}[0,1]$ shared across all modalities, $\mathbf{x}_t = (1-t)\mathbf{x}_0 + t\boldsymbol{\epsilon}$, $\boldsymbol{\epsilon} \sim \mathcal{N}(\mathbf{0}, \mathbf{I})$, with velocity target $\mathbf{v} = \boldsymbol{\epsilon} - \mathbf{x}_0$. For actions, this interpolation is applied in the normalized action space before the learned action projection. The training loss is
\begin{equation}
\begin{aligned}
    \mathcal{L}(\theta) \;=\;&
        \mathbb{E}_{\mathbf{x}_0,\, \boldsymbol{\epsilon},\, t}\!\Bigl[
            \lambda_f \bigl\lVert \hat{\mathbf{v}}^f - \mathbf{v}^f \bigr\rVert_2^2
            + \lambda_a \bigl\lVert \hat{\mathbf{v}}^a - \mathbf{v}^a \bigr\rVert_2^2
            \\[-1mm]
            &\qquad\qquad +
            \lambda_v \bigl\lVert \hat{\mathbf{v}}^v - \mathbf{v}^v \bigr\rVert_2^2
        \Bigr],
\end{aligned}
    \label{eq:loss}
\end{equation}
where $\lambda_a = 0$ for samples without action labels. By default, our model predicts motion and action only ($\lambda_f = 1$, $\lambda_a = 0.1$, $\lambda_v = 0$); the video term ($\lambda_v = 1$) is enabled only for ablations that predict future video. Actions use a 20-dimensional bimanual end-effector space (per arm: 3D position, 6D rotation, 1D gripper), normalized via 2nd/98th percentile clipping following~\cite{Barreiros2026Careful}.

\subsection{Training Procedure}
\label{sec:training_procedure}

We train our model in two stages using a mixture of egocentric human video and robot interaction data. Both stages use AdamW with a constant learning rate and bf16 mixed precision. Mid-training runs ${\approx}14{,}000$ optimizer steps at global batch size $512$ and learning rate $2.8\times10^{-4}$. Fine-tuning runs ${\approx}1{,}500$ steps at batch size $256$ and learning rate $5.6\times10^{-5}$.

\textbf{Stage 1: Mid-training.}
We train on the full data mixture combining human video and robot interaction datasets. Human video samples are supervised on the scene-flow loss only ($\lambda_a = 0$, $\lambda_v = 0$), while robot samples are supervised on scene flow and action. Dataset-specific batch weights control the relative sampling frequency. This stage exposes the backbone to diverse egocentric motion at scale, building a strong motion prior before any action-specific finetuning.

\textbf{Stage 2: Fine-tuning.}
In a second stage, we exclusively train on robot interaction data, dropping the human video stream entirely. This sharpens action prediction without diluting the gradient signal with action-free samples, and we raise the action loss weight to $\lambda_a = 1$.

\subsection{Inference}
\label{sec:inference}

At test time, given an initial observation $\mathbf{o}_0$ and language
instruction $\ell$, our model jointly denoises its output heads by integrating the learned velocity field with the UniPC ODE solver~\cite{Zhao2023UniPC} under a shifted noise schedule (shift $5$) for $N = 10$ steps; we apply no classifier-free guidance. The robot executes the full predicted chunk of $T = 16$ actions before re-planning from the next observation. One cycle takes ${\approx}0.7$\,s on a single NVIDIA H100 GPU, and asynchronous execution overlaps planning with robot motion.

\section{Experiments}
\label{sec:experiments}
We organize our experiments for \modelLogo{} around three questions.

\textbf{(1) Does predicting 3D motion rather than video improve policy performance?}
We compare \modelName{} with matched video-prediction variants and prior methods on in-distribution real-robot tasks, out-of-distribution real-robot tasks, and the RoboTwin benchmark (Secs.~\ref{sec:in-dist}--\ref{sec:robotwin-single-task}).

\textbf{(2) Does action-free human video enable out-of-distribution transfer?}
We remove the human-video portion of the training mixture and measure the resulting change in performance on novel objects, tasks, and motion primitives (Sec.~\ref{sec:human-video}).

\textbf{(3) Does motion serve an effective intermediate for action prediction?}
We use an inverse-dynamics probe to measure how efficiently actions can be recovered from ground-truth motion versus ground-truth video, and analyze how the action stream allocates attention across modalities (Secs.~\ref{sec:id-scaling} and~\ref{sec:attn-mask}).

\subsection{Experimental Setup}
\label{sec:setup}

\textbf{Datasets.}
We train on a mixture of action-free egocentric human video and action-labeled bimanual YAM robot teleoperation. The human-video data is primarily drawn from \textbf{EgoVerse}~\cite{Punamiya2026EgoVerse}: $81$\,h of egocentric human video, filtered for diversity from a $1{,}362$-hour corpus. The robot component comprises $10$\,h selected for diversity from $700$ hours of \textbf{MolmoAct2-BimanualYAM} teleoperation~\cite{Fang2026MolmoAct2}, $3.7$\,h of internal task-directed YAM demonstrations, and $0.4$\,h of internal undirected YAM play. The physical YAM setup copies that of MolmoAct2. We also include $0.7$\,h of internal, action-free human demonstrations on tasks that overlap with robot data. Human clips provide video and scene-flow supervision. Robot demonstrations also provide action supervision. Mid-training uses the full mixture. Fine-tuning uses the in-house YAM demonstration and play data.

\textbf{Baselines.}
We select baselines for each experiment according to each evaluation's main question. 
For in-distribution real-world evaluation, we compare \modelName{} with matched variants that substitute joint motion-action (MA) prediction for video-action (VA) prediction and video-motion-action (VMA) prediction to isolate whether action prediction benefits from motion, video, or both.

For out-of-distribution real-robot evaluation, we additionally compare against two prior works that represent close alternatives to our method: \textbf{AMPLIFY}~\cite{Collins2025AMPLIFY}, which predicts sparse, tokenized 2D point tracks as an intermediate between video and action, and \textbf{DreamZero}~\cite{Ye2026DreamZero}, which predicts future video jointly with actions. This comparison tests whether 3D motion transfers better than either image-space motion or pixel prediction under shifts in objects, tasks, and motion primitives. We train all the models on the data mixture described above until convergence.

On the RoboTwin \cite{chen2025robotwin} benchmark, we compare against $\bm{\pi_{0.5}}$~\cite{Black2025Pi05}, \textbf{FAST-WAM}~\cite{Yuan2026FastWAM}, and an action-only \textbf{Diffusion Policy} \cite{chi2025diffusion} (Behavior Cloning) baseline. These choices represent a pretrained Vision-Language-Action (VLA) policy, a video-based WAM, and a policy trained without any video pretraining. The Behavior Cloning model is a 20M parameter, 8-layer diffusion transformer with the Wan architecture. For Fast-WAM, we use the same rank-64 Wan-backbone LoRA configuration as \modelName{} and match the trainable capacity of its action head. We fine-tune $\pi_{0.5}$ end-to-end using its native training recipe, evaluating the VLA in its standard adaptation setting. Across methods, we hold the data and evaluation protocol fixed.  

\textbf{Evaluation protocol.}
For real-world rollouts, we report task success rate (SR), as the fraction of rollouts in which the policy completes the task. For the out-of-distribution evaluation, we also report task progress (TP), as the average fraction of task stages completed. We conduct $15$ rollouts per task for the in-distribution real-robot evaluation and $10$ rollouts per task for the out-of-distribution evaluation. The no-human-video ablation is done over $5$ rollouts per task. In RoboTwin, we report success in $100$ Easy-setting episodes per task and method.

\subsection{Policy Evaluation}
\begin{figure}[!t]
    \centering
    \vspace*{2mm}  
    \setlength{\abovecaptionskip}{0pt}
    \caption{\textbf{In-distribution success rates.}
    Across 15 rollouts FloMo performs best on every task; the matched video--motion--action
    (VMA) and video--action (VA) variants lower success.}
    \label{fig:policy-indist}
    \vspace{3pt}
    \includegraphics[width=\linewidth]{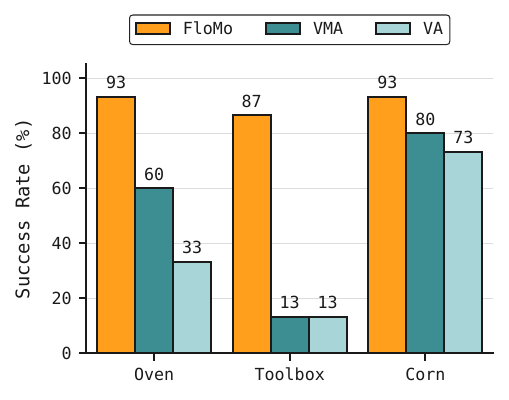}
\end{figure}

\begin{table}[!t]
\centering
\setlength{\tabcolsep}{1.5pt}
\footnotesize
\caption{\textbf{Out-of-distribution success rates and human-video ablation.}
We evaluate generalization across unseen motion primitives, tasks, and objects.
The main comparison uses $10$ rollouts per condition ($30$ total); the
no-human-video ablation uses $5$ rollouts per condition ($15$ total).
SR = successful rollouts / total; TP = avg.\ fraction of subtasks completed (\%).}
\label{tab:policy-ood}
\vspace{2pt}

\begin{tabular}{@{}lcccccccc@{}}
\toprule
\multirow{2}{*}{Method}
    & \multicolumn{2}{c}{Close Cabinet}
    & \multicolumn{2}{c}{Pull String}
    & \multicolumn{2}{c}{Highlighter}
    & \multicolumn{2}{c}{Aggregate} \\
\cmidrule(lr){2-3}
\cmidrule(lr){4-5}
\cmidrule(lr){6-7}
\cmidrule(lr){8-9}
 & SR & TP & SR & TP & SR & TP & SR & TP \\
\midrule

VMA
    & $0/10$ & $0$
    & $0/10$ & $0$
    & $2/10$ & $45$
    & $2/30$ & $15$ \\

VA
    & $0/10$ & $20$
    & $1/10$ & $10$
    & $1/10$ & $15$
    & $2/30$ & $15$ \\

DreamZero
    & $5/10$ & \textbf{75}
    & $1/10$ & $15$
    & $0/10$ & $0$
    & $6/30$ & $30$ \\

AMPLIFY
    & \textbf{6/10} & $70$
    & $0/10$ & $0$
    & $0/10$ & $0$
    & $6/30$ & $23$ \\

\modelLogo{} (no human video)
    & $0/5$ & $20$
    & $1/5$ & $50$
    & $0/5$ & $20$
    & $1/15$ & $30$ \\

\midrule

\textbf{\modelLogo{} (MA)}
    & \textbf{6/10} & $70$
    & \textbf{7/10} & \textbf{70}
    & \textbf{5/10} & \textbf{65}
    & \textbf{18/30} & \textbf{68} \\

\bottomrule
\end{tabular}
\end{table}

\subsubsection{In-Distribution Real-Robot Tasks}\leavevmode\par\nobreak
\label{sec:in-dist}

\textbf{Setup.}
We evaluate on three contact-rich, precision tasks on the bimanual YAM platform, modelling that from MolmoAct2~\cite{Fang2026MolmoAct2}. \emph{Oven} requires opening a toaster-oven door, retrieving a croissant from inside, and placing it on a plate. \emph{Toolbox} requires opening a hinged lid via a small crevice, grasping a screwdriver, and placing it in the container. \emph{Corn} requires grasping a corn cob and placing it in a bowl. We run $15$ rollouts per task and report success rate. We compare \modelName{} against the VMA and VA prediction-target variants.

\textbf{Results.}
Figure~\ref{fig:policy-indist} reports in-distribution results. \modelName{} achieves the highest success rate on every task, with an average success rate of $91\%$, compared with $51\%$ for VMA and $40\%$ for VA. Under the matched architecture and training protocol, adding video prediction to the motion--action objective reduces average success by $40$ percentage-points, while replacing motion prediction with video prediction reduces it by $51$ percentage-points. The difference is largest in \emph{Toolbox}, where \modelName{} achieves $87\%$ success while both video-prediction variants achieve only $13\%$. Opening the toolbox requires locating a narrow crevice and excuting contact-rich, precise motion. Predicting 3D motion directly emphasizes the geometry of this contact.

\begin{figure*}[!t]
    \centering
    \includegraphics[width=\textwidth]{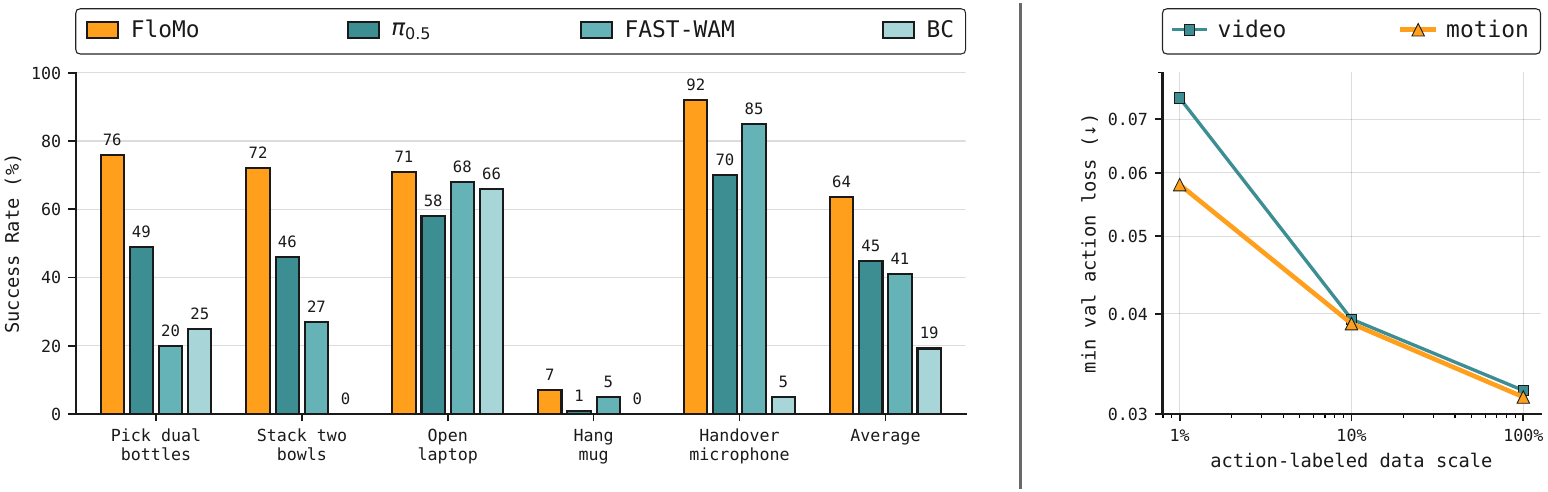}

    \caption{
    \textbf{(a) RoboTwin single-task evaluation.}
    Recorded RoboTwin Easy success rates across five tasks, with 100 evaluations per task and method. Across all tasks, FloMo performs the best.
    \textbf{(b) Inverse-dynamics scaling.}
    Held-out action error versus action-data scale for two \emph{ground-truth} conditioning signals (log-log).
    Recovering actions from motion is more sample-efficient than from video: the gap is largest at $1\%$ of the data and narrows as data grows.
    }
    \label{fig:robotwin-and-id}
\end{figure*}

\subsubsection{Out-of-Distribution Real-Robot Generalization}\leavevmode\par\nobreak
\label{sec:ood}

\textbf{Setup.}
We probe generalization along three axes, each isolating a distinct distribution shift. \emph{Close Cabinet} involves pushing a cabinet door shut, an unseen motion primitive in robot data during training. \emph{Pull String} requires pulling a hanging string to a target mark, an unseen task. \emph{Highlighter} introduces an unseen object (highlighter) which must be placed in a bowl. We run $10$ rollouts per condition and report success rate (SR) and task progress (TP) for \modelName{}, VMA, VA, AMPLIFY~\cite{Collins2025AMPLIFY}, and DreamZero~\cite{Ye2026DreamZero}. The no-human-video ablation in \ref{sec:human-video} uses $5$ rollouts per condition. Each task has a single intermediate stage: contacting the cabinet door for \emph{Close Cabinet}, grasping the string for \emph{Pull String}, and picking up the highlighter for \emph{Highlighter}. TP is the fraction of these two stages a rollout completes, averaged over rollouts: $0$ if it never reaches the intermediate stage, $50\%$ if it reaches it but does not finish, and $100\%$ if it completes the task.

\textbf{Results.}
Table~\ref{tab:policy-ood} reports out-of-distribution results. Among the full-data methods, \modelName{} is the only method that generalizes consistently across all three distribution shifts. It achieves an average success rate of $60\%$ and average task progress of $68\%$, compared with $20\%$ SR and $23\%$ TP for AMPLIFY, $20\%$ SR and $30\%$ TP for DreamZero, and $7\%$ SR and $15\%$ TP for both VMA and VA.

All methods that predict pixels through a video objective (VMA, VA, DreamZero), or through image-space 2D tracks (AMPLIFY), degrade in performance. We hypothesize that \modelName{} retains performance because its 3D scene-flow representation is invariant to appearance and camera movement in pretraining and mid-training data. Removing the image-space prediction target recovers generalization, and adding the video target, even alongside motion, coincided with lower OOD success in these evaluations.

We attribute this gap to what each target encodes. A video objective binds the model to scene-specific appearance, texture, and object identity, none of which are preserved under the distribution shifts we test. One possible explanation is sensitivity to training-set appearance statistics. By construction, 3D scene flow discards these factors and retains only the geometric transition between states, a representation that is invariant to appearance and shared across embodiments. This is what allows \modelName{} to transfer a motion primitive such as string-pulling, observed only in human video, to the robot: on \emph{Pull String}, \modelName{} succeeds on $7/10$ rollouts while no baseline exceeds $1/10$.

\subsubsection{RoboTwin Benchmark}\leavevmode\par\nobreak
\label{sec:robotwin-single-task}

\textbf{Setup.}
We evaluate single-task adaptation on five RoboTwin tasks spanning bimanual grasping (\emph{Pick Dual Bottles}), sequential placement (\emph{Stack Bowls}), articulated-object manipulation (\emph{Open Laptop}), precision placement (\emph{Hanging Mug}), and bimanual transfer (\emph{Handover Microphone}). For each task and method, we train a policy on the $50$ official demonstrations. The runs use $5{,}000$ gradient steps and a batch size of $256$.  We evaluate the models on $100$ episodes from the RoboTwin Easy setting.

\textbf{Results.}
Figure~\ref{fig:robotwin-and-id}(a) shows that \modelName{} achieves the highest success rate on all five robotwin tasks. Averaged across tasks, \modelName{} achieves $63.6\%$ success, compared with $44.8\%$ for $\pi_{0.5}$, $41.0\%$ for FAST-WAM, and $19.2\%$ for BC. These results show that predicting 3D motion provides an effective interface for transferring video pretraining to action prediction across a distinct set of simulated tasks.

\subsection{Transfer from Human Video}
\label{sec:human-video}

Table~\ref{tab:policy-ood} reports both the full-data comparison and the no-human-video ablation. To determine the impact of human video, we train a separate \modelName{} variant with the ${\approx}81$\,h EgoVerse subset and $0.7$\,h of internal action-free human demonstrations removed. Removing human video lowers average success from $60\%$ to $7\%$, suggesting that much of \modelName{}'s out-of-distribution performance comes from human video rather than robot demonstrations.

The per-subtask scores show where the model trained without human video fails. It can begin each task but rarely completes the novel motion: it contacts the cabinet on $2$ of $5$ trials but never closes it ($0/5$), grasps the string on $4$ of $5$ trials but successfully pulls it on only one, and grasps the highlighter on $2$ of $5$ trials but never places it in the bowl.

We verified this by direct text search of the training data. The robot data, both the MolmoAct2 episodes and the YAM demonstrations, contains no references to cabinets, strings, or markers. The EgoVerse subset, by contrast, contains all three motion classes explicitly: cabinet closing (e.g.\ ``close cabinet door under sink''), string pulling (e.g.\ ``pull string from spool''), and slender-object placement (e.g.\ ``place pen lid on pen, place pen in bowl''), even though the exact word ``highlighter'' never appears. Since these motions are reachable only through the human video, the transfer in Table~\ref{tab:policy-ood} likely originates there.

\subsection{Analysis of Motion as an Action Intermediate}

\subsubsection{Inverse-Dynamics Sample Efficiency}\leavevmode\par\nobreak
\label{sec:id-scaling}
The policy results above establish motion as the more effective prediction target for WAMs, but the source of this gain remains ambiguous: it could stem from more accurate forward-dynamics predictions or a more decodable conditioning representation for inverse dynamics. We isolate the latter with a controlled inverse-dynamics probe. We repurpose \modelName{} to predict future action chunks from \emph{ground-truth} future signals, eliminating forward-dynamics error so that any remaining difference in action error is attributable to the conditioning representation alone. We compare conditioning on ground-truth future scene flow against ground-truth future RGB, training each on $\{1\%, 10\%, 100\%\}$ of a filtered subset of \textbf{MolmoAct2-BimanualYAM} and reporting the minimum held-out action error.


Fig.~\ref{fig:robotwin-and-id}(b) demonstrates that motion is the more sample-efficient signal: at $1\%$ of the action data, it yields roughly $22\%$ lower action error than video ($0.058$ vs.\ $0.074$); the gap collapses to roughly $1\%-2\%$ at scales above $10\%$ of the dataset. Because the ground truth future is supplied in both conditions, this gap is a property of the representation quality rather than downstream performance, as in practice the actions are \textit{jointly predicted} with the video or motion rather than conditioned on ground truth futures. This experiment shows, however, that motion exposes the action-relevant structure directly without losing predictive information, whereas recovering it from raw RGB video takes more examples. Motion is thus both an easier target to predict and the more efficient representation of the information needed to predict actions.


\subsubsection{Action Attention Across Modalities}\leavevmode\par\nobreak
\label{sec:attn-mask}

To compare the importance of each modality for action prediction, we investigate the attention maps of the VMA variant.
We hypothesize that if motion is an effective intermediate in our evaluated setting, then action tokens should attend predominantly to motion tokens rather than video tokens, and this pattern should persist across embodiments.

\begin{figure}[!t]
\centering
\includegraphics[width=0.96\linewidth]{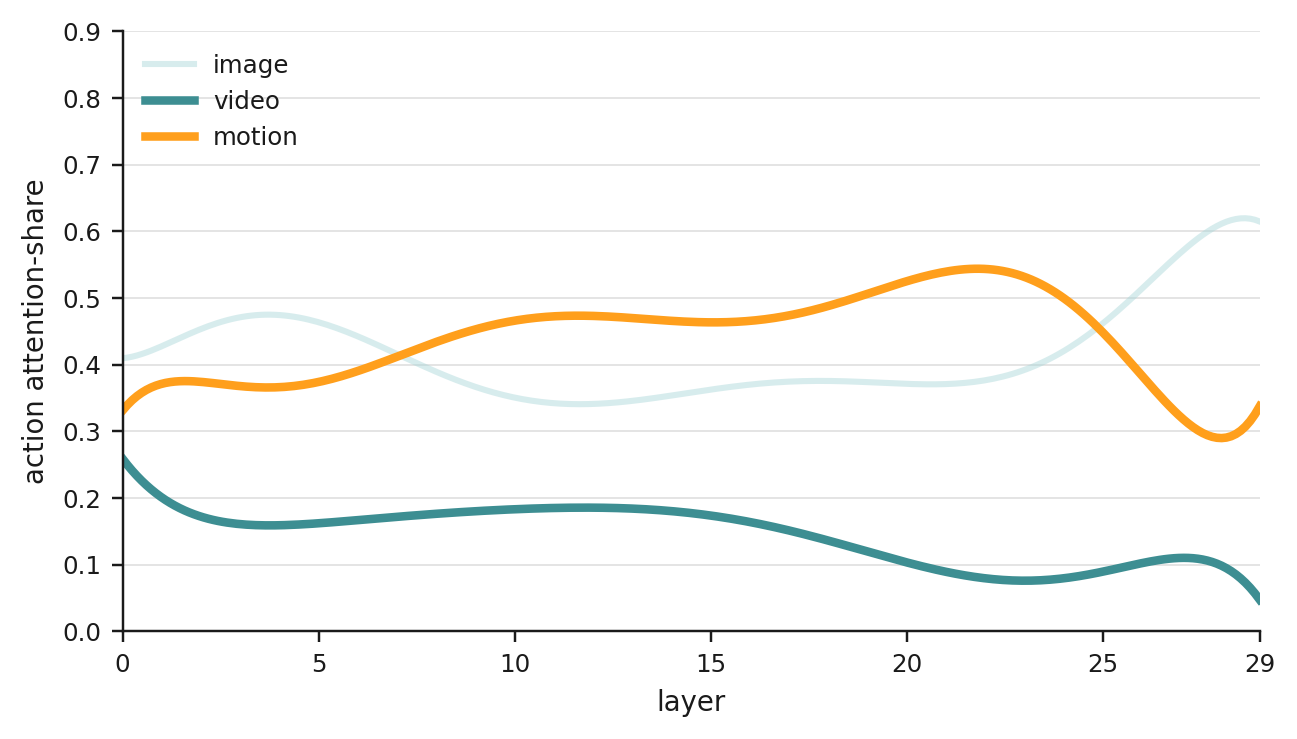}
\caption{\textbf{Action attention concentrates on motion tokens.} Share of action-query attention to each input modality, averaged over attention heads and robot validation rollouts, as a function of DiT layer. Attention to motion rises through the trunk and dominates at the late-middle layers, while attention to video stays lowest throughout; averaged over layers, action queries attend to motion roughly $3\times$ more than to video.}
\label{fig:attention-share}
\end{figure}

Figure~\ref{fig:attention-share} reports the share of action-query attention going to each input modality (image, video, motion) as a function of DiT layer, aggregated across attention heads, averaged over $1,000$ held-out samples. Beginning from an approximately uniform distribution at the input layer, action queries progressively reallocate attention toward motion tokens through the network trunk, reaching ${\sim}55\%$ at the late-middle layers. Attention to video exhibits the opposite trend, remaining the lowest across all layers and declining to ${\sim}10\%$ at the deepest layers. Averaged over attention heads and layers, action queries attend to motion approximately $3\times$ more than to video. Taken together, these results indicate that the action stream draws predominantly on motion rather than video, supporting our hypothesis that 3D scene flow is an effective intermediate in our evaluated setting.

\label{sec:discussion}




\section{Discussion and Limitations}
\label{sec:conclusion}


We argue that the priors of video pretraining are carried by a model's weights, not by what it is trained to predict. A world action model should target motion rather than pixels: an appearance-invariant geometric signal that an action head can decode directly. Rendering 3D scene flow into the latent space of a pretrained video generator lets a single backbone predict motion and action without a new tokenizer, inheriting internet-scale video priors through its weights. Across real-robot manipulation, predicting motion outperforms predicting video and jointly predicting video and motion, both in distribution and on unseen objects, motion primitives, and tasks; action queries attend far more to motion than to video; and a controlled inverse-dynamics probe shows motion is the more sample-efficient conditioning signal. These results point to motion as an effective intermediate in our evaluated setting.

Two limitations temper these results. First, extracting 3D scene flow adds an offline preprocessing step for every clip. Second, \modelName{} predicts motion and action jointly, and although the two usually agree, they sometimes diverge. A plausible predicted motion does not guarantee a correct action, because the action head is not constrained to be exactly consistent with the predicted flow, and distinct actions can in principle produce the same scene motion. In these cases, treating predicted motion as a proxy for the executed action breaks down, and tightening the coupling between the two predictions through mechanisms such as per-modality variance schedules or objectives to maximize mutual information between modalities is a natural direction for future work. Three additional failure modes occur: scene flow is predicted for the front camera only, so motion estimates degrade when the manipulated object is occluded in that view; instruction grounding is less reliable when a named object is unfamiliar or a similar distractor is present; and on occlusion-heavy clips the off-the-shelf tracker yields noisy trajectories that enter training as degraded targets.

%
%
%



\bibliographystyle{IEEEtran}
\bibliography{example}  

\clearpage

\appendices

\section{Implementation Details}
\label{appx:implementation}

\subsection{Architecture}
\modelName{} is initialized from the Wan~2.2 5B Text-Image-to-Video DiT~\cite{Wan2025}: a $30$-layer transformer with hidden width $d{=}3072$, $24$ attention heads (head dimension $128$), feed-forward width $14{,}336$, and a $(1,2,2)$ patchifier over the VAE latent grid. We use bidirectional (full) attention so every token attends to every other. On top of the shared backbone we add lightweight modality-specific heads:
\begin{itemize}[leftmargin=*,itemsep=2pt]
    \item \textbf{Action head.} Each per-timestep action vector $\mathbf{a}_i \in \mathbb{R}^{d_a}$ ($d_a{=}20$) is mapped to a DiT token by a single linear projection $\mathbb{R}^{d_a}\!\to\!\mathbb{R}^{d}$ plus a learned per-position embedding; predicted action tokens are read back out by a single linear projection $\mathbb{R}^{d}\!\to\!\mathbb{R}^{d_a}$.
    \item \textbf{Video and scene-flow heads.} Video frames and RGB-rendered scene-flow frames are both encoded and decoded by the frozen Wan~2.2 VAE. On the trunk side they reuse the backbone's own patchify (Conv3d) and unpatchify (linear) projections rather than adding new layers, so these heads contribute no trainable parameters.
    \item \textbf{Conditioning.} The initial observation $\mathbf{o}_0$ is encoded by the frozen Wan VAE into $\mathbf{z}^o$, which forms a clean prefix in the token sequence that is never noised; the instruction $\boldsymbol{\tau}$ is encoded by the frozen umT5-XXL text encoder~\cite{Chung2023UniMax} ($512$ tokens, width $4096$) and conditions the backbone through cross-attention. For the robot datasets the observation comprises three camera views (one front and two wrist), each VAE-encoded into its own prefix tokens; the human-video data uses a single view. Scene flow and video are predicted only in the front view.
\end{itemize}
We fine-tune the DiT with LoRA~\cite{Hu2022LoRA} ($r{=}64$, $\alpha{=}128$, dropout $0$), injecting adapters into the query/key/value/output projections of both self- and cross-attention and into both feed-forward linear layers of all $30$ blocks; the Wan VAE and umT5 stay frozen. This leaves ${\approx}161$M trainable parameters (LoRA adapters plus the small action head) on top of the frozen 5B backbone.

\subsection{Training Hyperparameters}
Both stages use AdamW ($\beta_1{=}0.95$, $\beta_2{=}0.999$, weight decay $10^{-6}$), gradient clipping at $1.0$, bf16 mixed precision, and a constant learning rate. All inputs are processed at $256{\times}256$ resolution, giving a $16{\times}16$ Wan-VAE latent grid, and each example spans a $16$-frame window: the model conditions on the initial frame and predicts the $16$-step action chunk together with the video and scene flow over the window. Training uses a flow-matching objective with $1{,}000$ timesteps. We augment every source except EgoVerse with random-resized-crop (scale $[0.85,1.0]$; ratio $[0.98,1.02]$) and color jitter (brightness, contrast, and saturation $0.25$; hue $0.1$). All models are fine-tuned from their respective epoch-$28$ mid-training checkpoints, using only in-house YAM data. Unless otherwise noted, fine-tuning evaluations use the epoch-$3$ checkpoint. Per-stage training settings are listed in Table~\ref{tab:appx-hparams}.

\begin{table}[H]
\centering
\small
\caption{\textbf{Per-stage training settings.}}
\label{tab:appx-hparams}
\vspace{5pt}
\begin{tabular}{lcc}
\toprule
 & \textbf{Mid-training} & \textbf{Fine-tuning} \\
\midrule
Learning rate (constant)               & $2.8\times10^{-4}$  & $5.6\times10^{-5}$ \\
Global batch size                      & $512$               & $256$ \\
Action loss weight $\lambda_a$         & $0.1$               & $1.0$ \\
Optimizer steps                        & ${\approx}14{,}000$ & ${\approx}1{,}500$ \\
\bottomrule
\end{tabular}
\end{table}

\begin{table*}[!t]
\centering
\small
\caption{\textbf{Dataset composition.} ``Tasks'' counts distinct task labels (EgoVerse uses fine-grained activity categories, the others coarse manipulation tasks); ``Clips'' counts clips, videos, or demonstrations (one recording each); ``Hours'' is the span of unique footage used in training: the number of distinct frames appearing in any $16$-frame window, divided by the source frame rate. All figures are for the subset actually used in each split.}
\label{tab:appx-data}
\vspace{7.5pt}
\begin{tabular}{l ccc ccc}
\toprule
 & \multicolumn{3}{c}{\textbf{Train}} & \multicolumn{3}{c}{\textbf{Validation}} \\
\cmidrule(lr){2-4}\cmidrule(lr){5-7}
\textbf{Dataset} & \textbf{Tasks} & \textbf{Hours} & \textbf{Clips} & \textbf{Tasks} & \textbf{Hours} & \textbf{Clips} \\
\midrule
EgoVerse              & $3{,}660$ & $81.0$ & $4{,}303$ & $35$ & $1.0$  & $46$  \\
MolmoAct2-BimanualYAM & $25$      & $10.0$ & $428$     & $22$ & $1.0$  & $46$  \\
In-house YAM          & $4$       & $3.7$  & $1{,}029$ & $4$  & $0.4$  & $113$ \\
In-house Human Data   & $11$      & $0.7$  & $710$     & $11$ & $0.1$  & $83$  \\
In-house Play YAM     & $162$     & $0.4$  & $173$     & $19$ & $0.05$ & $19$  \\
\bottomrule
\end{tabular}
\end{table*}

\subsection{Inference}
At test time, we draw the scene-flow and action tokens with a flow-matching sampler~\cite{Lipman2022FlowMatching}, integrating the learned velocity field with the UniPC ODE solver~\cite{Zhao2023UniPC} under a shifted noise schedule (shift $5$) for $N{=}10$ steps. We do not apply classifier-free guidance; the model is run on the conditional path only. We execute all $16$ predicted actions per planning cycle, re-planning from the next observation. Each planning cycle (one joint motion-and-action denoising of $N{=}10$ steps through the full 5B trunk) takes ${\approx}0.9$\,s on a single NVIDIA B200 GPU; asynchronous execution overlaps planning with robot motion.

\subsection{Loss Weighting and Modality Masking}
The training loss (Eq.~\ref{eq:loss}) uses $\lambda_f{=}1$, $\lambda_a{=}0.1$, and $\lambda_v{=}0$ by default; the video term ($\lambda_v{=}1$) is enabled only for video-prediction ablations. During fine-tuning (Stage~2) we raise the action weight to $\lambda_a{=}1$. Per-sample modality masking sets $\lambda_a{=}0$ for action-free samples (human video), allowing co-training across datasets with mismatched modality coverage.

\section{Dataset Details}
\label{appx:datasets}

\subsection{Source Datasets and Filtering}
Below we describe each source dataset and how it is filtered to the subset used in training; Table~\ref{tab:appx-data} summarizes the result.

\textbf{EgoVerse.} EgoVerse~\cite{Punamiya2026EgoVerse} is a large egocentric human-manipulation video dataset; from it we use an $828$\,h subset collected by Mecka AI. From this we build a task-balanced subset by round-robin over tasks (one clip per task, then a second clip per task, and so on), giving a $200$\,h manifest of $8{,}578$ clips over $3{,}696$ tasks. We hold out $9$ environments ($35$ tasks) for validation. From the training split we randomly subsample $16$-frame windows down to a fixed budget, shuffling first so the kept windows cover all tasks rather than clustering on a few. Mid-training uses this subset, which spans ${\approx}81$\,h of unique footage; no motion filtering is applied.

\textbf{MolmoAct2-BimanualYAM.} MolmoAct2-BimanualYAM~\cite{Fang2026MolmoAct2} is a bimanual-YAM robot teleoperation dataset ($737$ recordings, three $640{\times}360$ views at $30$\,fps). We drop $7$ ``spell-out'' tasks and round-robin across the remaining $25$ coarse tasks to a task-balanced selection, of which ${\sim}10$\,h ($428$ clips) is used for training.

\textbf{In-house YAM.} Teleoperated bimanual-YAM demonstrations of three tasks: put the corn in the bowl, open the oven and place the croissant on the plate, and open the toolbox and place the screwdriver inside. Each task includes randomized initial states as well as active data collection to enable data-efficient recovery from failure modes. This subset consists of $1{,}029$ demonstrations totaling ${\approx}3.7$\,h. Before training, data containing motion above or below pre-defined thresholds is filtered out.


\textbf{In-house Human Data.} Action-free human videos of the In-house YAM tasks, recorded from a single front-facing camera without action labels, so they supervise only the video and scene-flow heads. They cover the three robot tasks plus related variations (asparagus or pepper placed in a bowl or oven, pliers into the toolbox), for $710$ demonstrations (${\approx}0.7$\,h) across $12$ splits.

\textbf{In-house Play YAM.} A small set of unstructured play demonstrations on the YAM robot ($173$ demonstrations, ${\approx}0.4$\,h), filtered like the In-house YAM data and pooled as an auxiliary set during fine-tuning (Table~\ref{tab:appx-weights}).

\subsection{Scene-Flow Extraction}
We run SpatialTrackerV2~\cite{Xiao2025SpatialTrackerV2} on $T{=}16$-frame windows with $Q{=}4096$ query points uniformly arranged at $t{=}0$. We transform tracks into the current camera frame using estimated camera poses, then compute cumulative displacement (Eq.~\ref{eq:cumflow}). Extraction runs at roughly $2.4$ seconds per $16$-frame window on a single NVIDIA L40S GPU, or about $72$ GPU-hours per hour of $30$ FPS video. 

\subsection{Sampling Weights}
Table~\ref{tab:appx-weights} lists the weights that set how often each dataset is sampled in each training stage. Mid-training draws most samples from EgoVerse ($0.65$) to build a broad motion prior. It splits $0.25$ evenly between the two robot datasets, so action learning is supervised throughout, and gives the small in-house human and play sets $0.05$ each. Fine-tuning uses only action-labeled robot data: in-house YAM demonstrations ($0.75$) and in-house play ($0.25$).

\begin{table}[!t]
\centering
\small
\setlength{\tabcolsep}{5pt}
\caption{\textbf{Dataset sampling weights across training stages}.}
\label{tab:appx-weights}
\begin{tabular}{lcc}
\toprule
\textbf{Dataset} & \makecell{\textbf{Mid-training}\\\textbf{weight}} & \makecell{\textbf{Fine-tuning}\\\textbf{weight}} \\
\midrule
EgoVerse                 & 0.65  & 0.00  \\
MolmoAct2-BimanualYAM    & 0.125 & 0.00  \\
In-house YAM             & 0.125 & 0.75  \\
In-house Human Data      & 0.05  & 0.00  \\
In-house Play YAM        & 0.05  & 0.25  \\
\bottomrule
\end{tabular}
\end{table}

\end{document}